\documentclass[letterpaper]{article} 
\usepackage{aaai23}  
\usepackage{times}  
\usepackage{helvet}  
\usepackage{courier}  
\usepackage[hyphens]{url}  
\usepackage{graphicx} 
\usepackage{natbib}  
\usepackage{caption} 
\usepackage{multirow}
\usepackage{amsmath,cases}
\usepackage{booktabs}
\usepackage{bbding}
\usepackage{algorithm}
\usepackage{algorithmic}
\usepackage{eucal}
\usepackage{babel}[American]
\usepackage{newfloat}
\usepackage{listings}
\DeclareCaptionStyle{ruled}{labelfont=normalfont,labelsep=colon,strut=off} 
\floatstyle{ruled}
\newfloat{listing}{tb}{lst}{}
\floatname{listing}{Listing}
\usepackage{color}
\definecolor{mygreen}{RGB}{90,135,109}

\title{Infusing Definiteness into Randomness: \mbox{Rethinking Composition Styles for Deep Image Matting}
}
\author{
    Zixuan Ye\textsuperscript{\rm 1,$\dagger$} ~~~~
    Yutong Dai\textsuperscript{\rm 2,$\dagger$} ~~~~
    Chaoyi Hong\textsuperscript{\rm 1} ~~~~
    Zhiguo Cao\textsuperscript{\rm 1} ~~~~
    Hao Lu\textsuperscript{\rm 1,}\thanks{Corresponding author.\\
    \hspace*{4.5mm}$^\dagger$These authors contributed equally.}
}

\affiliations{
    \textsuperscript{\rm 1}
    School of Artificial Intelligence and Automation, Huazhong University of Science and Technology, China
    
    \textsuperscript{\rm 2}
    Australian Institute for Machine Learning, The University of Adelaide, Australia
    
    $\tt hlu@hust.edu.cn$
}

\usepackage{bibentry}

\begin{document}

\maketitle

\begin{abstract}

We study the composition style in deep image matting, a notion that characterizes a data generation flow on how to exploit limited foregrounds and random backgrounds to form a training dataset. 
Prior art executes this flow in a completely random manner by simply going through the foreground pool or by optionally combining two foregrounds before foreground-background composition. In this work, we first show that naive foreground combination can be problematic and therefore derive an alternative formulation to reasonably combine foregrounds. Our second contribution is an observation that matting performance can benefit from a certain occurrence frequency of combined foregrounds and their associated source foregrounds during training.
Inspired by this, we introduce a novel composition style that binds the source and combined foregrounds in a definite triplet. In addition, we also find that different orders of foreground combination lead to different foreground patterns, which further inspires a quadruplet-based composition style. Results under controlled experiments on four matting baselines show that our composition styles outperform existing ones and invite consistent performance improvement on both composited and real-world datasets. Code is available at:  \url{https://github.com/coconuthust/composition_styles}

\end{abstract}

\section{Introduction}

Large-scale datasets provide essential ingredients for training deep neural networks~\cite{ImageNet}. 
This sense also applies to deep image matting whose goal is to estimate the accurate alpha matte $\alpha$ that satisfies the matting equation
\begin{equation}
\label{eq:matting}
I = \alpha F+(1-\alpha) B\,,
\end{equation}
such that the foreground $F$ can be separated from the background $B$ given an image $I$. 
In particular, high-quality alpha mattes with accurate annotations are vital for training deep matting models.
However, 
large-scale matting datasets are difficult to collect. So far only a few datasets~\cite{DIM,HATT,AIM} with limited number of high-quality alpha mattes are publicly available. Therefore, how to maximize the value of limited alpha mattes to support the training of deep matting models has become a fundamental problem. 

\begin{figure}[!t]\small
  \centering
  \includegraphics[width=\linewidth]{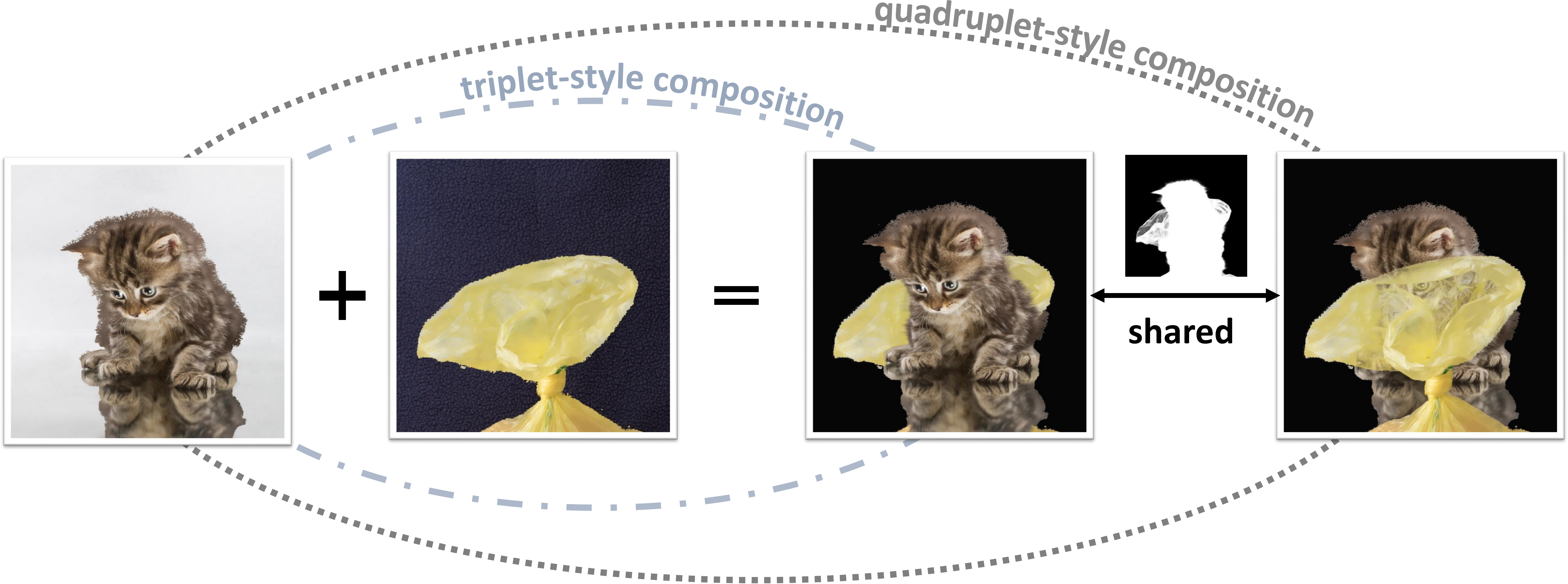}
  \vspace{-15pt}
  \caption{Two foregrounds can generate two additional combined foregrounds. Two source foregrounds and a combined one can construct a definite triplet. The combined one also has a twin combination of different foreground patterns, but shares the same alpha matte, which can be further used to form a quadruplet. They build the core of our proposed composition styles. 
  }
  \label{fig:mining_combined_foregrounds}
  \vspace{-10pt}
\end{figure}

\begin{figure}[!t]
  \centering
  \includegraphics[width=\linewidth]{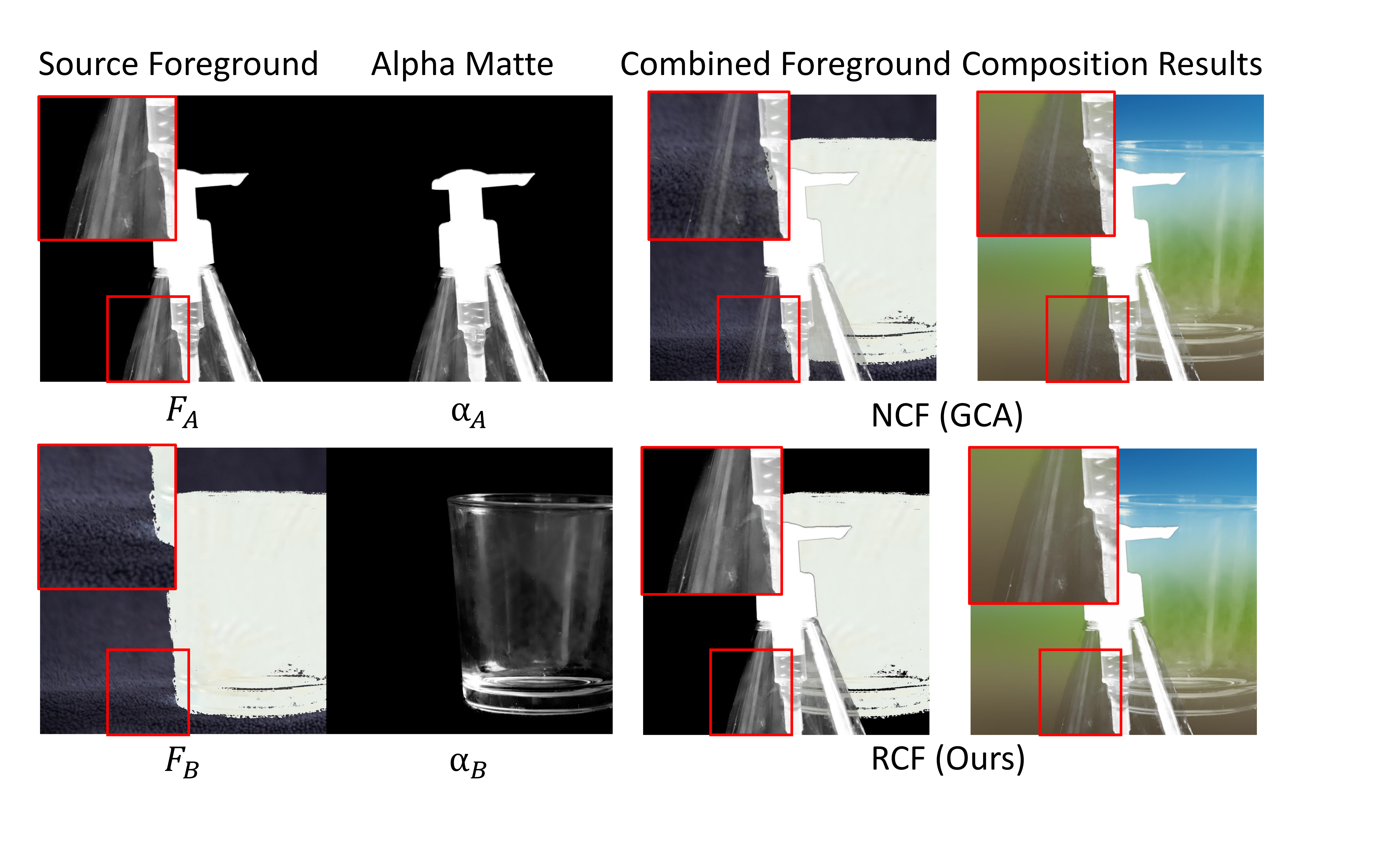}\vspace{-5pt}
  \caption{Combined foregrounds and composition results generated by two foreground combination operators $\mathcal{NCF}$ and our $\mathcal{RCF}$. $\mathcal{NCF}$ cannot filter out the noise in $F_B$, while $\mathcal{RCF}$ yields 
  a clean composition.}
  \label{fig:composition_results}
  \vspace{-5pt}
\end{figure}

A common strategy is to use image composition as in Eq.~\eqref{eq:matting} to synthesize samples. 
To form a training dataset, image composition is repetitively used following a data generation flow or a composition rule. For ease of exposition, we define such flows or rules as \textit{composition styles}, featured by foreground selection and foreground-background composition. In the open literature, two composition styles are widely used, say DIM-style composition~\cite{DIM} and GCA-style composition~\cite{GCA}. Thereinto, DIM-style composition iterates through the foreground pool; each time a chosen foreground is composited with a random background. 
GCA-style composition further
introduces an optional process of foreground combination~\cite{LBSampling} to generate the combined foreground 
(Fig.~\ref{fig:mining_combined_foregrounds}). To distinguish from foreground-background composition, we slightly abuse the term \textit{foreground combination} to indicate the composition between two foregrounds. Different from general-purpose augmentations that enhance appearance diversity, foreground combination boosts pattern diversity. However, we find that the naive combination of foregrounds ($\mathcal{NCF}$) used in GCA-style composition can be problematic. As shown in Fig.~\ref{fig:composition_results}, it fails to filter out the noise 
and generates artifacts in the new foreground. 

By probing 
$\mathcal{NCF}$, we find that it 
treats the second foreground as background and neglects its alpha matte. 
To solve this, we derive a new composition equation to achieve reasonable combination of foregrounds ($\mathcal{RCF}$).
In contrast to $\mathcal{NCF}$, $\mathcal{RCF}$ sequentially composites the two foregrounds onto the background. The new equation yields a different foreground representation and encodes the alpha mattes of both two foregrounds. Fig.~\ref{fig:composition_results} shows that $\mathcal{RCF}$ removes the artifacts in the combined foreground.

Further, we find the relation among source foregrounds and their combinations is underexplored in GCA-style composition; it executes foreground combination in a completely random manner. 
Such a process neglects what source foregrounds are used to form the combined foreground. Instead we consider foreground combination as a reversible chemical reaction where the source foregrounds and the combined one can mutually assist the learning of foreground patterns of each. Therefore, the occurrence frequency of a certain foreground should depend on the number of its combinations in the sample set. To meet this demand, we derive a composition style, termed \textit{triplet-style composition}, where relevant samples are bound in a triplet to ensure the positive correlation. In this way, the training set is formed by triplet groups.
Another interesting observation is that different combination orders of two foregrounds in $\mathcal{RCF}$ can result in two distinct combined foregrounds. They share the same alpha matte but differ in foreground patterns, which forms a twin relation. We therefore realize that 
we can extend foreground patterns by swapping the combination order of source foregrounds. 
We thus introduce \textit{quadruplet-style composition}, where a combined twin foreground is further included to form a quadruplet.

Extensive experiments under controlled conditions on four deep matting baselines, including IndexNet Matting~\cite{index}, GCA Matting~\cite{GCA}, A2U Matting~\cite{A2U} and MatteFormer~\cite{matteformer}, show that our composition styles indicate 
a clear advantage against previous composition styles, \textit{e.g.}, $12.7\%\sim17.9\%$ relative improvement in the gradient metric on IndexNet Matting. 
Our composition styles also show a good generalization on real-world images.
Moreover, we can achieve comparable performance against other composition styles even with half of the available foregrounds.

For the first time, we delve into the data generation flow in deep image matting and show that careful treatment of this flow can make a difference in matting performance.

\section{Related Work}

\subsubsection{Deep Image Matting.}\label{related_dim}
Image matting is an active research area that has yielded prolific improvements during the last decades~\cite{KNN,bayesian}. The emergence of deep learning further advances natural image matting. One milestone work is DIM~\cite{DIM}, which presents an end-to-end deep matting framework and collects a benchmark dataset. Enlightened by DIM, large bodies of follow-up work emerge.

Much existing effort has been made on improving network architectures for matting. GCA~\cite{GCA} designs a guided contextual attention module to propagate the opacity information based on low-level features.
To recover subtle details, IndexNet~\cite{index} and A2U~\cite{A2U} propose dynamic upsampling operators by predicting context-aware kernels. Recently, TIMI-Net~\cite{timi} attempts to mine the information within trimap with a specialized encoder. These works lift deep matting to a new height, while they seldom probe the potential wealth outside the network architecture.
Some work thinks outside the box and explores strong data augmentations. CA~\cite{CA} uses excessive data augmentations to improve the generalization of models. RMat~\cite{RMat} further improves data augmentations in CA to enhance the robustness of models on real-world images. In this work, we delve into the process before training begins, that is, how to form the training dataset with foreground and background pools. 

\begin{figure*}[!t]
  \centering
  \includegraphics[width=\linewidth]{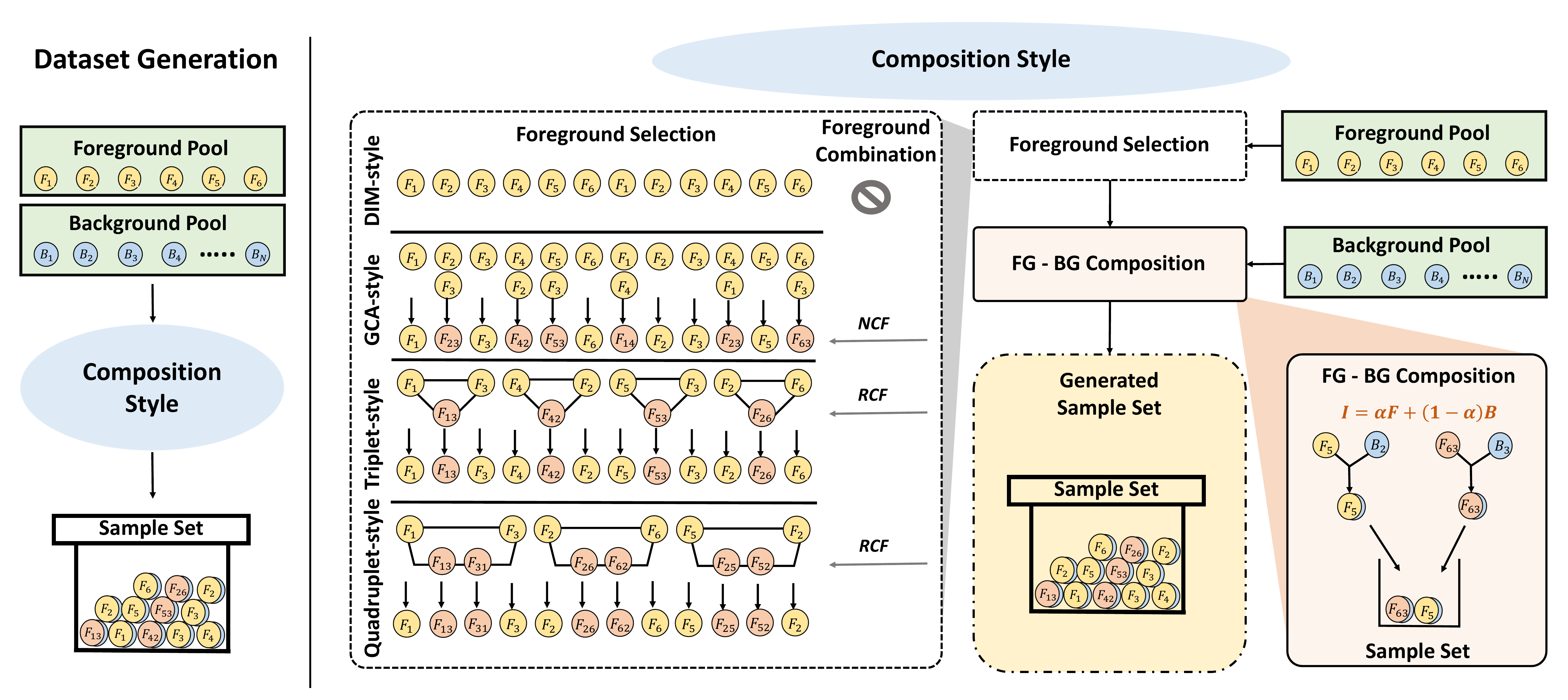} \vspace{-15pt}
  \caption{\textit{Left:} Pipeline of the data generation flow. \textit{Right:} The architecture of composition style and the comparison between previous composition styles and ours. Ours can establish a link between the source and combined foregrounds, and further increase the foreground pattern diversity brought by the source foregrounds.}
  \label{fig:styles}
\end{figure*}

\subsubsection{Data Augmentation.}
Data augmentation is 
widely-used 
in deep learning to increase data diversity. 
It can benefit both high-level tasks such as object detection~\cite{cutandpaste}, and low-level tasks such as image enhancement~\cite{density} and image matting~\cite{GCA,A2U}. They can be viewed as general-purpose techniques as no task-specific design is involved. Data augmentation is applied after the foreground selection and before the foreground-background composition. Therefore, the standard augmentations can also be add-ons to our composition styles.

In addition to standard augmentations, a stream of work adopts image composition to augment training data by leveraging its nature of generating synthetic images. For example, Mixup~\cite{mixup} 
and CutMix~\cite{cutmix} are adopted as data augmentations. Benefiting from the soft alpha mattes, the foreground patterns can be augmented by a combination of foregrounds based on image composition. However, we find that the treasure beneath the foreground combination remains unexplored in existing composition styles, so we delve into 
this research topic and particularly study the foreground selection behind foreground combination.

\section{A Recap of Composition Styles}
Due to the lack of diverse human-annotated alpha mattes, the majority of deep matting models are trained on composited images. While previous work typically follows a data generation flow to form a training dataset with synthetic compositions, such a flow has not been paid attention to in the literature. For the first time, we formally define this flow as the composition style and use this notion to depict how to use foregrounds, backgrounds, and alpha mattes to form a (training) sample set.

Assume that we need to generate $12$ samples 
from a foreground pool with $6$ foregrounds and a background pool with infinite backgrounds. 
The generation process can be decomposed into two sub-stages: foreground selection and foreground-background (FG-BG) composition. As in Fig~\ref{fig:styles}, foreground selection refers to how to determine the $12$ foregrounds that constitute the $12$ samples from the foreground pool, and FG-BG composition refers to the alpha blending (Eq.~\eqref{eq:matting}) between the each foreground and a random background, which generates the sample. 
Each sample will be placed into a sample set after FG-BG composition.
Since different composition styles share the same FG-BG composition step, the key that distinguishes different styles lies in the foreground selection. 

In open literature, two composition styles are commonly used, \textit{i.e.}, DIM-style composition~\cite{DIM} and GCA-style composition~\cite{GCA}. Before we present our proposition, we first revisit these two composition styles. 

\subsection{DIM-Style Composition} 
Being the first attempt that introduces a composited training set for deep matting, DIM-style composition~\cite{DIM} implements naive composition by compositing each foreground onto different random backgrounds. Concretely, it iterates through 
the foreground pool to 
harvest the required number of foregrounds. For instance in Fig~\ref{fig:styles}, 
it iterates from $F_1$ to $F_6$ twice to acquire $12$ foregrounds before generating the sample set with FG-BG composition. 

\subsection{GCA-Style Composition} 

Inspired by~\cite{LBSampling}, GCA-style composition~\cite{GCA} introduces an additional 
foreground combination step, 
where two foregrounds are combined to generate a new foreground with a probability of $p$. Formally, given two foregrounds $F_A$ and $F_B$ and their alpha mattes $\alpha_A$ and $\alpha_B$, a new foreground $F_{\mathrm{new}}$ and a new alpha matte $\alpha_{\mathrm{new}}$ can be generated using 
Eq.~\eqref{eq:matting}, which treats $F_A$ as the foreground and $F_B$ as the background. 
We define an operator $\mathcal{NCF}$ to characterize such naive combination of foregrounds
\begin{equation}\small
    (F_{\mathrm{new}},\alpha_{\mathrm{new}})\leftarrow\mathcal{NCF}(F_A,\alpha_A,F_B,\alpha_B)\,,
\end{equation}
where
\begin{align}\small
\label{eq:f_new_gca}
&F_{\mathrm{new }}=\alpha_{A}F_A + (1-\alpha_{A})F_B\,,\\
\label{eq:alpha_new_gca}
&\alpha_{\mathrm{new }}=1-(1-\alpha_{A})(1-\alpha_{B})\,.
\end{align}
Akin to DIM-style composition, GCA-style composition follows the same FG-BG composition stage. 
They differ in foreground selection, where GCA-style combines foregrounds randomly. It first chooses source foregrounds like DIM-style composition, \textit{i.e.}, 
iterating through the foreground pool. For each source foreground, it has a probability of $p=0.5$ to be combined with 
another random foreground in the foreground pool. As shown in Fig~\ref{fig:styles}, $6$ of the $12$ source foregrounds are combined with random foregrounds by $\mathcal{NCF}$.

\section{Proposed Composition Styles \label{sec:tri and quad}
}

While DIM- and GCA-style compositions are effective, we show that they have certain weaknesses and may lead to sub-optimal compositions. Here we first derive a reasonable foreground combination operator $\mathcal{RCF}$ and then present our proposed composition styles built on top of $\mathcal{RCF}$. 

\subsection{Reasonable Combination of Foregrounds}
\label{ssec:foreground combination}
As aforementioned, GCA-style composition blends two foregrounds $F_A$ and $F_B$ to generate a new foreground $F_{\mathrm{new}}$ and a new alpha $\alpha_{\mathrm{new}}$. We agree with the representation of $\alpha_{\mathrm{new}}$; however, we have a different opinion on $F_{\mathrm{new}}$, \textit{i.e.}, according to Eq.~\ref{eq:f_new_gca}, only $\alpha_A$ is active, while $\alpha_B$ is 
neglected. 
Hence, we rethink foreground combination from a two-stage perspective and derive an alternative 
formulation of $F_{\mathrm{new}}$ 
by considering both $\alpha_A$ and $\alpha_B$.

We first composite a foreground, say $F_B$, onto a background $B$ to form a 
temporary background $B_{\mathrm{tmp}}$ by
\begin{equation}\small
B_{\mathrm{tmp}}=\alpha_{B}F_{B} + (1-\alpha_B)B\,.
\end{equation}
At the second stage, we overlay the other foreground, say $F_A$, onto 
$B_{\mathrm{tmp}}$ to generate the composition $I_{\mathrm{new}}$ by
\begin{equation}\small
\begin{aligned}
\label{eq:I_new_ours}
I_{\mathrm{new}} &=\alpha_{A} F_{A}+\left(1-\alpha_{A}\right)B_{\mathrm{tmp}}\\
&=\alpha_{A} F_{A}+\left(1-\alpha_{A}\right) \alpha_{B} F_{B}+\left(1-\alpha_{A}\right)\left(1-\alpha_{B}\right) B\,.
\end{aligned}
\end{equation}
By matching Eq.~\eqref{eq:I_new_ours} with $I_{\mathrm {new }}=\alpha_{\mathrm {new }} F_{\mathrm {new }}+\left(1-\alpha_{\mathrm {new }}\right) B$, one can infer
\begin{align}\small
\label{eq:fgcom}
    \begin{cases}
    \alpha_{\mathrm {new}} F_{\mathrm {new }}&=\alpha_{A} F_{A}+(1-\alpha_{A}) \alpha_{B} F_{B}\,,\\
    1-\alpha_{\mathrm {new}}&=(1-\alpha_{A})(1-\alpha_{B}) \,.
    \end{cases}
\end{align}
Given Eq.~\eqref{eq:fgcom}, we define a new operator $\mathcal{RCF}$ to implement reasonable combination of foregrounds such that
\begin{equation}\small
    (F_{\mathrm{new}},\alpha_{\mathrm{new}})\leftarrow\mathcal{RCF}(F_A,\alpha_A,F_B,\alpha_B)\,,
\end{equation}
 where
\begin{align}\small
&F_{\mathrm {new }}=\frac{\alpha_{A} F_{A}+\left(1-\alpha_{A}\right) \alpha_{B} F_{B}}{\alpha_{\mathrm {new }}+\epsilon}\,,\\
&\alpha_{\mathrm {new }}=1-\left(1-\alpha_{A}\right)\left(1-\alpha_{B}\right)\,.
\end{align}
$\epsilon$ is a small number (\textit{e.g.}, 1e-6) used to prevent zero division. Compared with $\mathcal{NCF}$, 
the information of both $\alpha_A$ and $\alpha_B$ is encoded by $\mathcal{RCF}$. 
According to Fig.~\ref{fig:composition_results}, $\alpha_B$ can play a key role in filtering out noise in $F_B$. Without $\alpha_B$, the artifacts in $F_B$ will passed to the combined foreground and 
affects the composition quality (cf. the boxed region in Fig.~\ref{fig:composition_results}), which may also influence model learning.  

\begin{figure*}[!t]
  \centering
  \includegraphics[width=\linewidth]{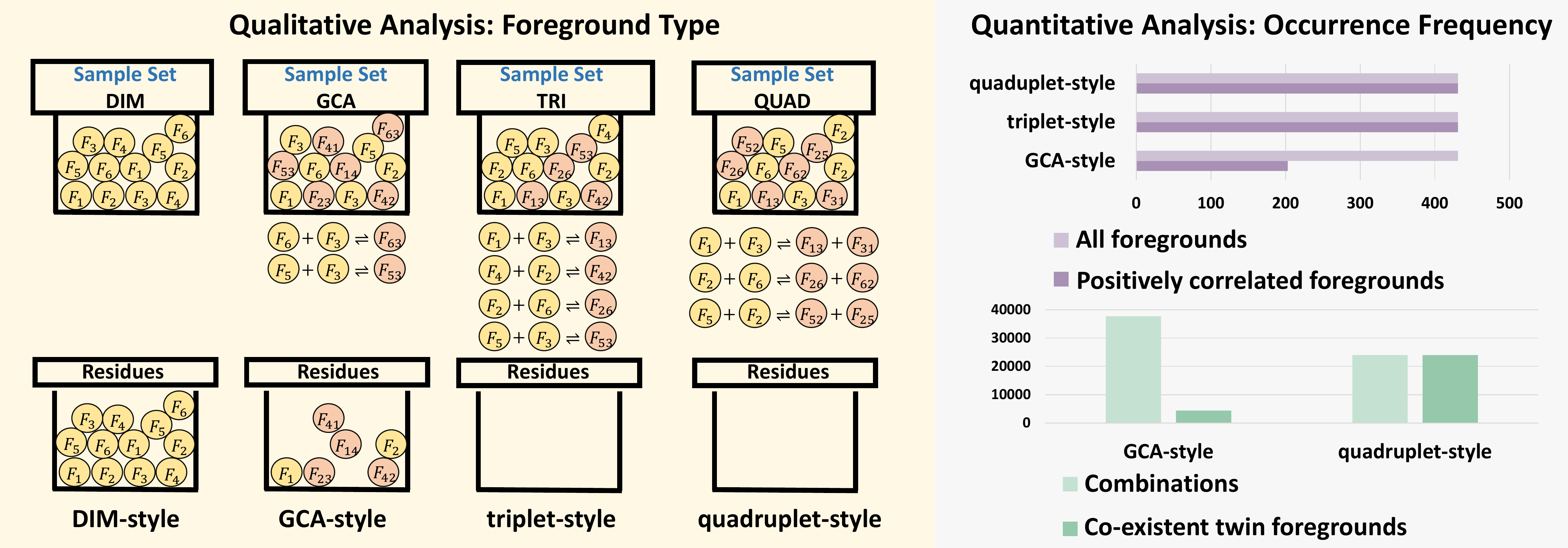} \vspace{-15pt}
  \caption{Foreground component analysis. \textit{Left:} Qualitative analysis of foreground type (useful vs. useless foregrounds). After removing the useful foregrounds in the reversible reaction, the useless foregrounds are left as residues. \textit{Right:} Quantitative analysis of occurrence frequency on i) the positively correlated foregrounds and ii) the co-existent twin foregrounds.}
  \label{fig:analysis}
\end{figure*}

\begin{figure}[!t]
  \centering
  \includegraphics[width=\linewidth]{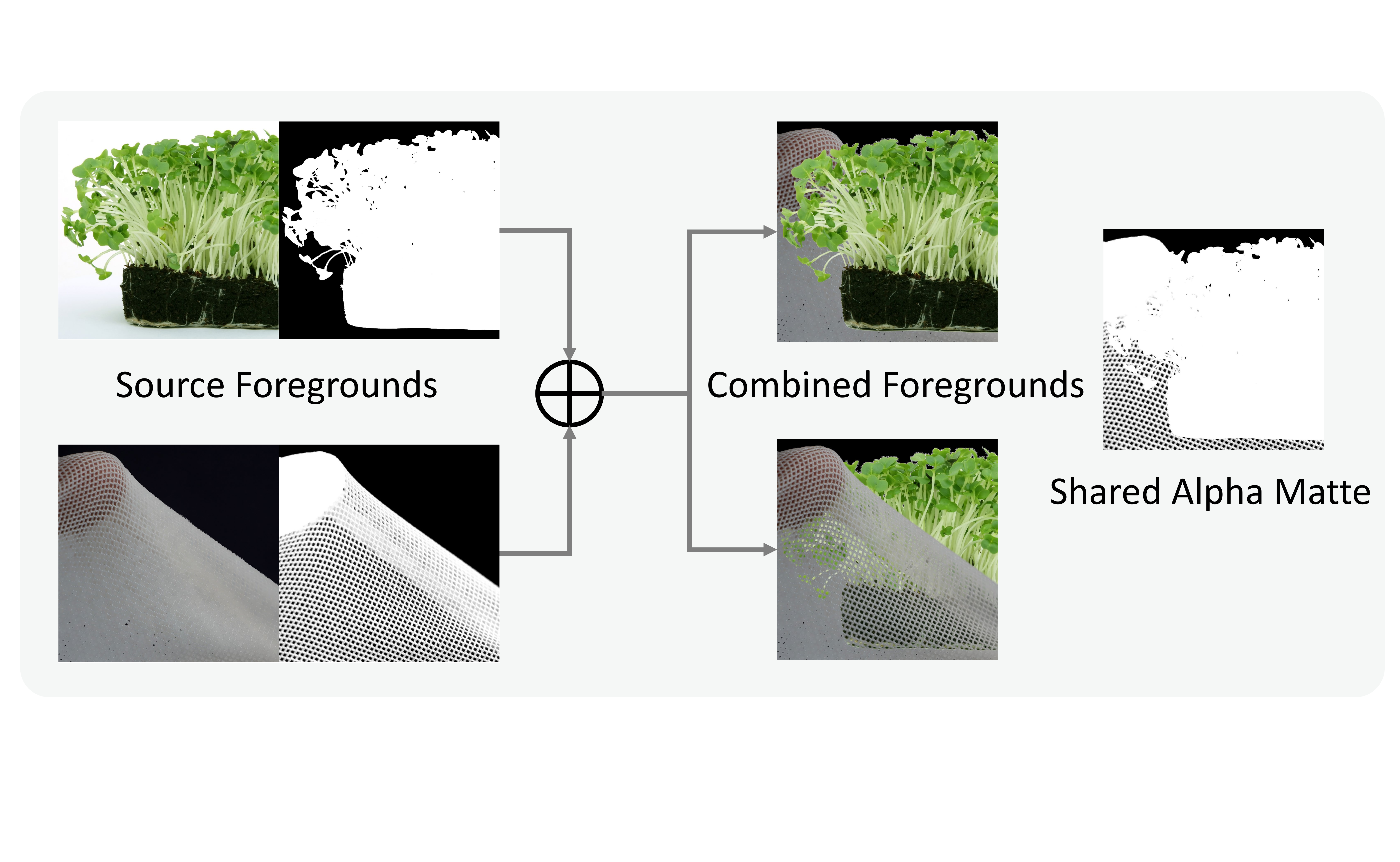} \vspace{-10pt}
  \caption{Combined foregrounds and alpha mattes in quadruplet-style composition.}
  \label{fig:shared}
\end{figure}

\subsection{Triplet-style Composition}
\label{triplet}
While $\mathcal{NCF}$ and $\mathcal{RCF}$ generate the new foreground, they neglect the prior knowledge of source foregrounds. For example, from Fig.~\ref{fig:composition_results}, the combined alpha matte exhibits new glass patterns; however, if a network does not see its source glass patterns sufficiently during training, it is likely to be confused in the inference of easy cases. To ease difficulties, we argue that the relation between the new foreground and the source foregrounds should be linked directly. 
Indeed, for the combined foreground, the source foregrounds can supplement the information that helps clarify the semantic meaning of complex patterns; for the source foregrounds, the inclusion of the combined foreground 
can assist the learning of combination rules. 
Therefore, 
by borrowing the concept of chemical reaction, we consider foreground combination as a reversible reaction
\begin{equation}\small
\label{eq:rever_tri}
F_{A}+F_{B} \rightleftharpoons F_{AB} \,.
\end{equation}
In this work, we view the source foregrounds and the combined foreground in the reaction above as \textit{useful foregrounds}.
Another important observation of this work is that, when the useful foregrounds appear with a certain proportion in the sample set, 
neither more nor less, the matting network can learn from foreground patterns more effectively and lead to better performance. 

However, previous work~\cite{LBSampling,GCA} does not consider the link and neglects the complementary relation between the source and combined foregrounds. That is, the sample set 
can contain many useless foregrounds. To re-link the relation, we devise a novel composition style, termed triplet-style composition, which binds the useful foregrounds in a definite triplet $(F_A,F_B,F_{AB})$. 
From Fig.~\ref{fig:styles}, the $12$ source foregrounds construct $4$ triplet groups. Samples formed by the triplet groups would have a stronger correlation than those formed by the unconstrained groups in GCA-style composition. 

\subsection{Quadruplet-Style Composition} 
\label{ssec:qd}
The triplet-style composition builds a link between parents (source foregrounds) and a child (the combined foreground). 
Interestingly, we find that 
the child actually has a twin brother, \textit{i.e.}, a Siamese combination. This is inspired by an observation that, if one swaps the combination order of two source foregrounds, 
the combinations can be different, \textit{i.e.},
\begin{equation*}\small
\label{assert}
F_{A} \text { overlay } F_{B} \neq F_{B} \text { overlay } F_{A}\,.
\end{equation*}
This claim can be easily validated by proof of contradiction.
Assume ``$\small F_{A} \text { overlay } F_{B} = F_{B} \text { overlay } F_{A}$'', then according to the definition of $F_{\text{new}}$ in Eq.~\eqref{eq:fgcom}, we have 
\begin{equation*}\small
\frac{\alpha_{A} F_{A}+\left(1-\alpha_{A}\right) \alpha_{B} F_{B}}{1-\left(1-\alpha_{A}\right)\left(1-\alpha_{B}\right)+\epsilon} =\frac{\alpha_{B} F_{B}+\left(1-\alpha_{B}\right) \alpha_{A} F_{A}}{1-\left(1-\alpha_{A}\right)\left(1-\alpha_{B}\right)+\epsilon}\,.
\end{equation*}
Since $\alpha_{\mathrm{new}}=1-\left(1-\alpha_{A}\right)\left(1-\alpha_{B}\right)$ remains unchanged, $F_A$ and $F_B$ must
meet the following condition 
\begin{equation*}\small
\alpha_{A}\alpha_{B} F_{B}=\alpha_{B} \alpha_{A} F_{A} \,.
\end{equation*}
It is clear that our assumption is valid if and only if $F_A =F_B$. Hence, the order matters. Another point is that, while the combined twin foregrounds are different, they share the same alpha matte, just like twins who are similar in most ways but still differ in certain aspects. This can also be observed in Fig.~\ref{fig:shared}. 
We therefore introduce the quadruplet-style composition, where an additional product has been added into the reversible reaction of the triplet-style composition (cf. Eq.~\eqref{eq:rever_tri}) as
\begin{equation}\small
\label{eq:rever_quad}
F_{A}+F_{B} \rightleftharpoons F_{AB} + F_{BA}\,.
\end{equation}
Compared with triplet-style composition, the only difference is that quadruplet-style composition generates from a foreground pair a quadruplet $(F_A, F_B, F_{AB}, F_{BA})$. From Fig.~\ref{fig:styles}, the resulting foreground selection is formed by $3$ quadruplet groups in quadruplet-style composition.



\subsection{Analysis of Different Composition Styles}\label{sec:comparison}
DIM-style composition increases sample diversity by varying backgrounds, but 
does not extend foreground patterns.
GCA-style composition further 
expands foreground diversity by foreground combination. They, however, fail to i) model the relation between the combined and source foregrounds and ii) mine the treasure beneath the two source foregrounds.
We address the issues above 
via two novel composition styles. In particular, triplet-style composition establishes the link between three relevant foregrounds, while GCA-style composition does not have such an explicit link. To show their differences, 
we present \textit{foreground component analysis}.
We analyze the foreground component from two aspects: qualitative analysis on foreground type and quantitative analysis on occurrence frequency.
The former distinguishes between useful and useless foregrounds based on the contribution of a certain foreground to the training of the whole sample set. 
The latter provides an empirical measure.
Note that we ignore the influence of the background, because i) the background pool can have infinite backgrounds and ii) matting models mainly 
learn foreground patterns.

As shown in Fig.~\ref{fig:analysis}, four sample sets generated by four composition styles have different foreground components. For example, the sample set generated by DIM-style composition has no combined foreground. If removing foregrounds involved in reversible reactions, the remaining residues are the useless foregrounds. Compared with the useful ones which assist mutual learning of foreground patterns, the useless ones are only limited to learning of their own patterns. The empty residue of the sample sets generated by proposed composition styles indicate that all foregrounds are useful and contribute to effective learning of other foreground patterns. 

\begin{table*}[!t] \footnotesize
\renewcommand\arraystretch{1.2}
\resizebox{\linewidth}{!}{
\begin{tabular}{@{}lcccccccccccccccc@{}}
\toprule
                 & \multicolumn{4}{c}{IndexNet Matting} & \multicolumn{4}{c}{GCA Matting}        & \multicolumn{4}{c}{A2U Matting}        &
                 \multicolumn{4}{c}{MatteFormer}     \\
                 \cmidrule(r){2-5} \cmidrule(r){6-9} \cmidrule(r){10-13} \cmidrule(r){14-17}
                 & SAD    & MSE &Grad & Conn & SAD   & MSE &Grad &Conn    & SAD   & MSE  &Grad & Conn  & SAD & MSE& Grad &Conn \\
\midrule
Reported         
& 45.80 &0.0130  & 25.90 & 43.70    
& 35.28 & 0.0091 & 16.92 & 32.53     
& 32.10 & 0.0078 & 16.33 & 29.00 
&23.80  &0.0041  &8.72   &18.89\\
DIM-style        
&43.03  &0.0115     &22.21 & 41.70     
&37.74  &0.0100     &18.64 &32.64       
&36.16  &0.0089     &17.20 &34.06 
&29.29  &0.0063     &10.85 &24.77\\
GCA-style        
&45.19  &0.0130  &23.62  &44.68      
&35.16  &0.0090  &17.69  &32.61 
&32.15  &0.0082  &16.39  &29.25
&23.71  &0.0039  &8.57   &20.47\\
Triplet-style    
&\underline{38.65}&\underline{0.0100}&\underline{20.29} &\underline{36.78}      
&\textbf{31.91}   &\textbf{0.0075}   &\textbf{14.14}    & \textbf{27.77} 
&\underline{31.52}&\underline{0.0072}&\underline{15.89} & \underline{28.47}
&\underline{23.26}&\textbf{0.0038}   &\textbf{7.34}     &\underline{19.52}\\
Quadruplet-style
&\textbf{38.16}   &\textbf{0.0099}   &\textbf{19.37}    &\textbf{36.31}      &\underline{33.16}&\underline{0.0080}&\underline{15.09} & \underline{29.35} 
&\textbf{30.47}   &\textbf{0.0071}   &\textbf{15.30}    & \textbf{27.47}
&\textbf{22.60}   &\textbf{0.0038}   &\underline{7.67}  &\textbf{19.26}\\
\bottomrule
\end{tabular}
}
\vspace{-5pt}
\caption{Quantitative comparisons of composition styles across $4$ baselines on the Composition-1K dataset. Best results are in boldface, and second-best ones are underlined. 
The 'Reported' row cites the published results for reference. The original IndexNet use the DIM-style composition, and the others use the GCA-style composition.}
\label{tab:styles}

\end{table*}

In addition to the residues, we can also understand from the occurrence frequency of foreground. The occurrence frequency $N$ of a single foreground should relate to that of the combinations it relates to. 
Formally, they should be positively correlated, \textit{i.e.},
\begin{equation}\small
N(F_A)\propto N(F_{AX}+F_{XA})\,,
\end{equation}
where $F_X$ is another foreground that is combined with 
$F_A$. As shown in Fig.~\ref{fig:analysis}, we count how many of the foregrounds follows the positive correlation; the results show that only about half of the foregrounds in GCA-style composition meet the requirement. However, our triplet-style composition allows all the foregrounds to have positive correlations by ensuring
\begin{equation}\small
N(F_A) = N(F_{AX}+F_{XA})\,,
\end{equation}
while our quadruplet-style composition by 
\begin{equation}\small
2N(F_A) = N(F_{AX}+F_{XA})\,.
\end{equation}

To highlight the difference between quadruplet-style composition and GCA-style composition, we also count the occurrence frequency of co-existence of twin foregrounds. The statistics in Fig.~\ref{fig:analysis} show that only $11.6\%$ of the combinations in GCA-style composition own the opportunity to have their twin foregrounds, 
while our quadruplet-style composition 
boosts such a proportion to $1$. Therefore, the quadruplet-style composition ensures that, once a combined foreground appears, its twin foreground will definitely appear in the sample set. 

\section{Results and Discussion}
To verify the effectiveness of the proposed two composition styles and the foreground combination operator $\mathcal{RCF}$, we conduct extensive experiments over different matting baselines on several benchmark datasets. We evaluate on both the synthetic and real-world scenarios to validate the generalization capability.

\subsection{Implementation Details}
Our implementation is based on $\tt{PyTorch}$. We only need to modify the dataloader 
without any architecture change when applying the composition styles.

\paragraph{Datasets.} 
We train models on the synthetic Adobe Image Matting~\cite{DIM} dataset and report performance on both the synthetic Composition-1K~\cite{DIM} dataset and the real-world AIM-$500$~\cite{AIM} and PPM-$100$~\cite{MODnet} datasets.  

\paragraph{Evaluation Metrics.} 
Following~\cite{metrics}, we employ four standard metrics to evaluate the predicted alpha mattes, including sum of absolute differences (SAD), mean squared errors (MSE), gradient errors (Grad), and connectivity errors (Conn).

\paragraph{Baselines.} 
We evaluate on four state-of-the-art matting baselines: IndexNet Matting~\cite{index}, GCA Matting~\cite{GCA}, A2U Matting~\cite{A2U}, and MatteFormer~\cite{matteformer}. 

\paragraph{Fairness.}
All experiments are conducted under the following controlled settings to fairly compare different composition styles.
First, we apply the same data augmentation strategies used in GCA Matting for different composition styles, and in this way the composition style is the only variable. 
Second, to eliminate the performance differences caused by different data volumes, we provide each composition style with exactly the same amount of training samples. Furthermore, we fix the order and the visual contents of the sample set when comparing the two foreground combination operators $\mathcal{NCF}$ and $\mathcal{RCF}$, to ensure that the only difference is the generation manner of $F_\mathrm{new}$.

\begin{table}[!t]
\renewcommand\arraystretch{1.3}
\resizebox{\linewidth}{!}{
\begin{tabular}{@{}lcccccccc@{}}
\toprule
& \multicolumn{4}{c}{AIM-500} & \multicolumn{4}{c}{PPM-100}\\
\cmidrule(r){2-5} \cmidrule(r){6-9}
& SAD    & MSE  & Grad   & Conn     & SAD   & MSE    & Grad & Conn \\
\midrule
DIM-style
&27.28 &0.0305 &21.37 &15.08 
&44.58 &0.0264 &41.83 &35.35 \\
GCA-style
&28.39&\underline{0.0303}&22.47 &14.71 
&41.01 &0.0249 &41.15 &31.87\\
Triplet-style
&\textbf{26.33} & \textbf{0.0278}&\textbf{19.61} &\textbf{12.90} 
&\textbf{38.94} &\underline{0.0234} &\textbf{38.64} &\underline{30.21}\\
Quadruplet-style
&\underline{26.65} &0.0307 &\underline{20.79} &\underline{14.37}
&\underline{40.22} &\textbf{0.0226} &\underline{40.19} &\textbf{30.02}\\
\bottomrule
\end{tabular}
}
\vspace{-5pt}
\caption{Quantitative comparisons of composition styles on real-world datasets with MatteFormer as the baseline.}
\label{tab:generalize}
\end{table}

\begin{figure*}[!t]
  \centering
  \includegraphics[width=\linewidth]{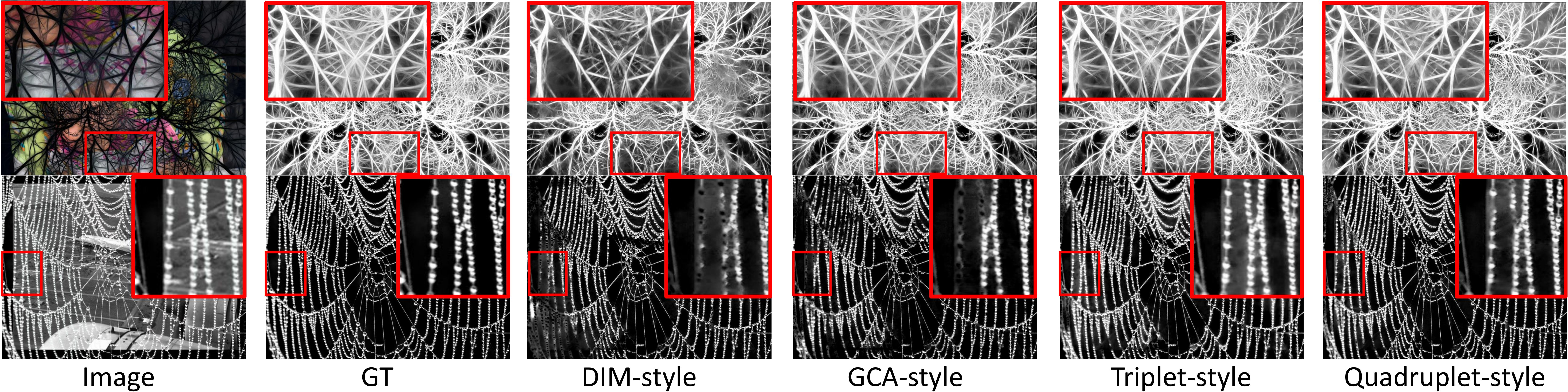}\vspace{-5pt}
  \caption{Qualitative results of different composition styles on Composition-1K dataset. 
  Due to the infused definiteness in our composition styles, ours produces better visualizations on complex textures.
  }
  \label{fig:visual_results}
  \vspace{-5pt}
\end{figure*}

\subsection{Comparison of Composition Styles}
We compare the composition styles from three aspects: effectiveness on the composited dataset, generalization ability on real-world datasets, and robustness to the number of foregrounds.

\paragraph{Effectiveness.}
We first compare the triplet- and quadruplet-style composition with prior composition styles on Composition-1K. 
From Table~\ref{tab:styles}, both triplet- and quadruplet-style compositions consistently outperform the DIM-style and the GCA-style compositions on all baselines, suggesting that our composition styles are generic designs. 
Establishing a definite link reveals the value of the foreground combination, with a significant $6.54$ SAD reduction on the IndexNet Matting baseline. In the open literature, such improvements often come from increased network complexity or an elaborate training strategy. However, we 
achieve this with only an improved data generation flow prior to model training,
without any change of network design or 
complicated training procedures.
Visual results are shown in Fig.~\ref{fig:visual_results}. Compared with the previous styles, our proposed ones can recover more clear, detailed structure such as complex wire patterns.

\paragraph{Generalization.}
Since the composition styles are developed for composited data, it is important to know the generalization of the styles on real-world data. 
Here MatteFormer is chosen as the baseline, and we report performance on two real-world datasets in Table~\ref{tab:generalize}. Results show that models trained with our composition styles also outperform other composition styles on read-world data. 

\begin{figure}[!t]
  \centering
  \includegraphics[width=\linewidth]{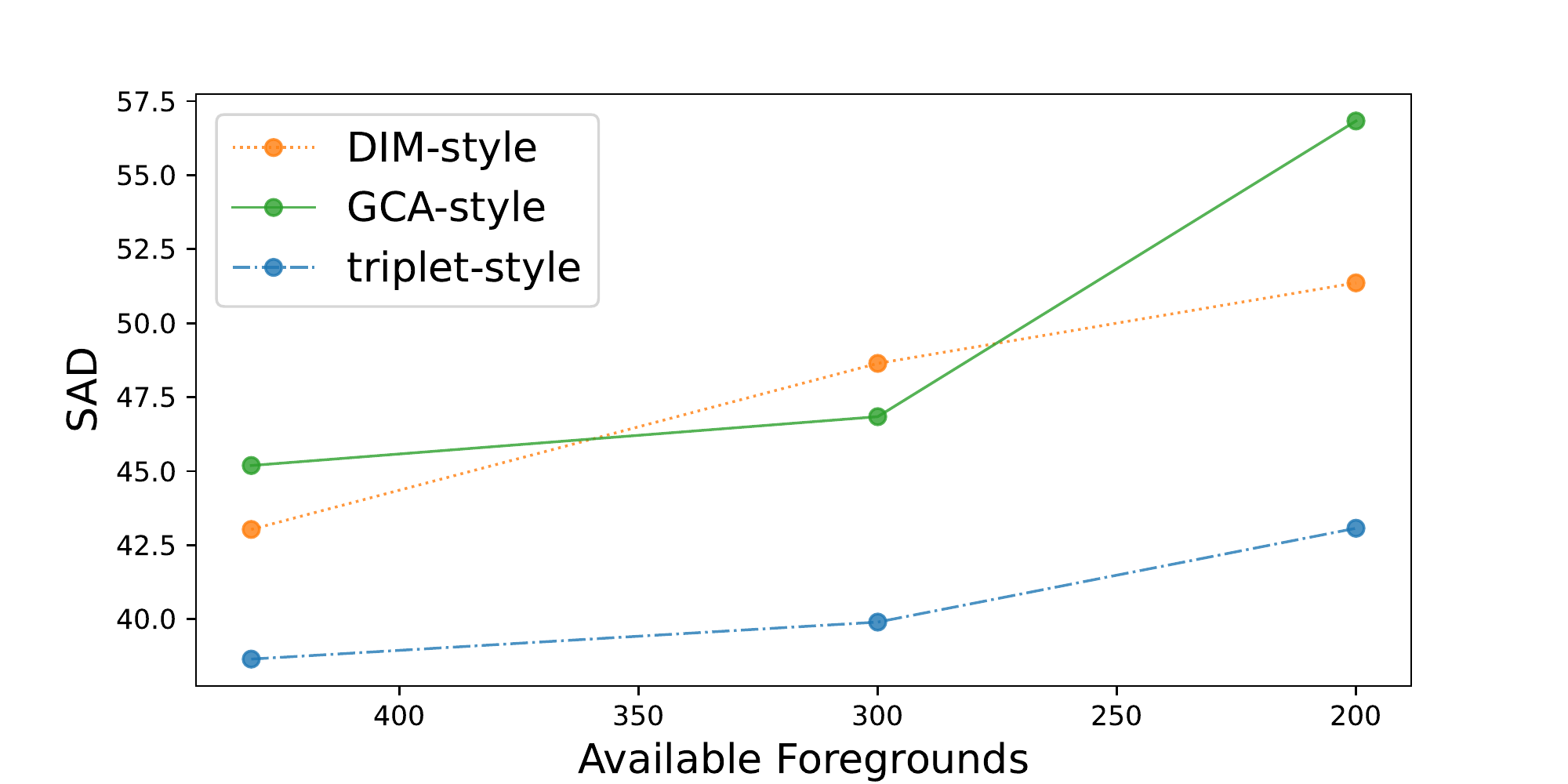}\vspace{-5pt}
  \caption{Robustness to the amount of available foregrounds. 
  IndexNet is used as the baseline. 
  }
  \label{fig:varying_foregrounds}
\end{figure}

\paragraph{Robustness.}
We also explore the robustness of different styles with reduced amount of available foregrounds. In Fig.~\ref{fig:varying_foregrounds}, the triplet-style is more robust than other styles when the available foregrounds decrease. Despite the foregrounds are halved, we still achieve comparable performance with the DIM-style that uses double amount of foregrounds.

\subsection{Comparison of Foreground Combination}
Here we compare $\mathcal{NCF}$ with the proposed $\mathcal{RCF}$ on A2U with the quadruplet-style and on IndexNet with GCA-style as the baselines, to show that the operator is independent of specific composition styles and matting models. We report performance on Composition-1K in Table~\ref{tab:foreground-combination}. Results show that $\mathcal{RCF}$ 
outperforms $\mathcal{NCF}$ on both baselines, confirming the effectiveness and the fundamental role of $\mathcal{RCF}$. 


\subsection{Ablation Study of Data Preprocessing Components}

We use 
IndexNet without any data processing as the baseline (D1). To validate the effectiveness of each component, the ablation study incorporates 
common data augmentation, the proposed 
$\mathcal{RCF}$, and the triplet-/quadruplet-style composition.
Results are reported in Table~\ref{tab:component}. By comparing D1 with D2, one can see an incremental improvement of $1.84$ SAD achieved by varying the visual appearance. By comparing D2 with D3 and D4, 
the proposed 
composition styles yield a substantial performance improvement.
Such results corroborate the effectiveness of our composition styles and the importance of a balanced foreground distribution. 
In addition, there is also a 
marginal improvement when the triplet style is altered to the quadruplet style (D3 vs.\ D4).


\begin{table}[!t]\small
\centering
\addtolength{\tabcolsep}{0pt}
\resizebox{\linewidth}{!}{
\begin{tabular}{@{}lcccccc@{}}
\toprule
 & \multirow{2}{*}{$\mathcal{NCF}$} & \multirow{2}{*}{$\mathcal{RCF}$}& \multicolumn{4}{c}{Compostion-1k} \\
 & && SAD & MSE & Grad & Conn \\
 \midrule
A2U & \Checkmark &  & 32.70 & \textbf{0.0066} & 16.24 & 30.29 \\
A2U &  & \Checkmark & \textbf{30.47} & 0.0071 & \textbf{15.30} & \textbf{27.27} \\
IndexNet & \Checkmark &  &45.19  &\textbf{0.0130}  &\textbf{23.62}  &44.68  \\
IndexNet &  & \Checkmark & \textbf{44.03} & 0.0135 & 24.76 & \textbf{42.89}\\
\bottomrule
\end{tabular}
}
\vspace{-5pt}
\caption{Comparison of foreground combination operator on the Composition-1K. A2U and IndexNet are adopted as baselines, respectively.}
\label{tab:foreground-combination}
\end{table}

\begin{table}[!t]\small
\addtolength{\tabcolsep}{-2pt}
\centering
\resizebox{\linewidth}{!}{
\begin{tabular}{@{}lcccccccc@{}}
\toprule
&DA  & $\mathcal{RCF}$ & Tri & Quad & SAD & MSE & Grad & Conn \\
\midrule
D1 & &  &    &  & 44.87 & 0.0124 & 25.08 & 41.23 \\
D2 &\Checkmark   &  &  &  & 43.03 & 0.0115 & 22.21 & 41.70 \\
D3 &\Checkmark   & \Checkmark & \Checkmark &  & 38.65 & 0.0100 & 20.29 & 36.78 \\
D4 &\Checkmark   & \Checkmark &  & \Checkmark & 38.16 & 0.0099 & 19.37 & 36.31\\
\bottomrule
\end{tabular}
}
\vspace{-5pt}
\caption{Ablation study of different components. 
`DA' denotes the data augmentation such as random jittering used in GCA Matting. `Tri' and `Quad' denote triplet- and quadruplet-style composition, respectively.}
\label{tab:component}
\vspace{-10pt}
\end{table}

\section{Conclusion}
In this work, we introduce the concept of the composition style to characterize the data generation flow in deep image matting. 
We first present an improved operator $\mathcal{RCF}$ to reasonably combine foregrounds with reduced artifacts. We then propose a triplet-style composition and a quadruplet-style composition, which infuses definiteness into the random data generation flow. 
We validate our propositions over four state-of-the-art matting baselines on both composited and real-world datasets. Results show that our composition styles consistently outperform the previous ones and demonstrate good generalization on real scenes. 

Being the first work that delves into the data generation flow in deep image matting, we believe our work points a good direction for addressing deep matting with improved composition styles. 

\vspace{5pt}
\noindent\textbf{Acknowledgement.} This work is supported by the National Natural Science Foundation of China under Grant No. 62106080.
\bibliography{aaai23}

\newpage

\renewcommand\thesection{S\arabic{section}}
\renewcommand\thefigure{S\arabic{figure}}

\section{Appendix}
We provide the following contents in this appendix:
\begin{itemize}
    \item[-] The implementation of four composition styles summarized in algorithm;
    \item[-] Output visualizations of different composition styles on real-world images;
    \item[-] Output visualizations of different foreground combination operators;
    \item[-] More results and ablation studies.
\end{itemize}

\section{Algorithm Form of Four Composition Styles}
Here we summarize the four composition styles in form of algorithms.
\subsection{A Table of Notations}
For ease of understanding, we define the notions as the Tab.~\ref{tab:notations}.
\begin{table}[h]
  \caption{Table of notations}
  \label{tab:notations}
  \begin{tabular}{cc}
    \toprule
    Notation & Description \\
    \midrule
    $\mathcal{S}$ & Sample set \\
    $\mathcal{F}$ & Foreground pool \\
    $\mathcal{B}$ & Background pool \\
    $\mathcal{A}$ & Alpha matte pool\\
    \midrule
    $F_X$ & Single foreground \\
    $\alpha_X$ & Single foreground\\
    $B_X$ & Single background\\
    $F_{AB}$ & The combination of $F_{A}$ and $F_{B}$\\
    \midrule
    $\mathcal{NCF}$ & Naive combination operator of foregrounds\\
    $\mathcal{RCF}$ & Reasonable combination operator of foregrounds\\
    $\mathcal{COMP}$ & FG-BG composition operator\\
  \bottomrule
\end{tabular}
\end{table}

\subsection{DIM-style Composition}
DIM-style composition selects foregrounds by iterating the foreground pool then perform FG-BG composition on the selected foregrounds. Specifically, for each composition it blends a foreground $F\in\mathcal{F}$ with a random background $B\in\mathcal{B}$ using the alpha matte $\alpha$ to form a sample $S$ such that $S\leftarrow \mathcal{COMP}(F, \alpha, B)$, where $\mathcal{COMP}$ is an operator that implements alpha blending. We summarize the data generation flow in Algorithm~\ref{alg:dim-style}.
\begin{algorithm}[h] \small
\caption{DIM-Style Composition} 
\label{alg:dim-style} 
\begin{algorithmic}
\REQUIRE Foreground set $\mathcal{F}$ with $N$ foregrounds, alpha set $\mathcal{A}$, and background set $\mathcal{B}$
\ENSURE Sample set $\mathcal{S}$ with $M$ samples
\STATE Initialize $\mathcal{S}\leftarrow\{\}$
\FOR {$i=1,...,M$}
    \STATE randomly choose $B\in\mathcal{B}$
    \STATE select $(F, \alpha) = (F_{i\%N}, \alpha_{i\%N}$)
    \STATE $S\leftarrow \mathcal{COMP}(F,\alpha,B)$
    \STATE $\mathcal{S}\leftarrow \mathcal{S} \cup \{S\}$ \textcolor{mygreen}{// include $S$ into $\mathcal{S}$}
\ENDFOR
\end{algorithmic}
\end{algorithm}

\subsection{GCA-style Composition}
Similar to DIM-style composition, GCA-style chooses basic foregrounds in iteration and provides a probability of $0.5$ to perform foreground combination. GCA-style composition is summarized in Algorithm~\ref{alg:gca-style}.
\begin{algorithm}[h] \small
\caption{GCA-Style Composition} 
\label{alg:gca-style} 
\begin{algorithmic}
\REQUIRE Foreground set $\mathcal{F}$ with $N$ foregrounds, alpha set $\mathcal{A}$, and background set $\mathcal{B}$
\ENSURE Sample set $\mathcal{S}$ with $M$ samples
\STATE Initialize $\mathcal{S}\leftarrow\{\}$
\FOR {$i=1,...,M$}
    \STATE randomly choose $B\in\mathcal{B}$
    \STATE select $(F_1, \alpha_1) = (F_{i\%N}, \alpha_{i\%N}$)
    \STATE Randomize $p\in[0,1]$
    \STATE \textcolor{mygreen}{// foreground combination with a probability of $0.5$}
    \IF{$p<0.5$} 
    \STATE randomly choose $(F_2,\alpha_2) \in (\mathcal{F},\mathcal{A})$
    \STATE $(F_{\mathrm{new}},\alpha_{\mathrm{new}})\leftarrow \mathcal{NCF}(F_1,\alpha_1,F_2,\alpha_2)$
    \textcolor{mygreen}{// combine $F_1$ with $F_2$}
    \STATE $S\leftarrow \mathcal{COMP}(F_{\mathrm{new}},\alpha_{\mathrm{new}},B)$ 
    \ELSE   
    \STATE $S\leftarrow \mathcal{COMP}(F_1,\alpha_1,B)$
    \ENDIF
    \STATE $\mathcal{S}\leftarrow \mathcal{S} \cup \{S\}$ \textcolor{mygreen}{// include $S$ into $\mathcal{S}$}
\ENDFOR
\end{algorithmic}
\end{algorithm}

\subsection{Triplet-style Composition}
Triplet-style composition need to generate an index pool $\mathcal{I}$ first, which refers to all possible two-by-two index combinations in the foreground library. Suggest the foreground pool contains $N$ foregrounds, then $\mathcal{I}=[(1,2),(1,3)...(N-1,N)]$. Each time we choose a pair of indexes $(m,n)$ from $\mathcal{I}$, then $(F_1, \alpha_1) = (F_m, \alpha_m)$ and $(F_2, \alpha_2) = (F_n, \alpha_n)$. Given $(F_1, \alpha_1)$, $(F_2, \alpha_2)$, and random backgrounds, we first generate two samples conditioned on source foregrounds. Then we generate the combined foreground and alpha matte using the $\mathcal{RCF}$ operator such that $(F_{\mathrm{new}},\alpha_{\mathrm{new}})\leftarrow \mathcal{RCF}(F_1,\alpha_1,F_2,\alpha_2)$. Triplet-style composition is summarized in Algorithm~\ref{alg:tri-style}.

\begin{algorithm}[h]\small
\caption{Triplet-Style Composition} 
\label{alg:tri-style} 
\begin{algorithmic}
\REQUIRE Foreground set $\mathcal{F}$, alpha set $\mathcal{A}$, and background set $\mathcal{B}$
\ENSURE Sample set $\mathcal{S}$ with $M$ samples
\STATE Initialize $\mathcal{S}\leftarrow\{\}$
\STATE \textcolor{mygreen}{// each iteration generates 3 samples}
\FOR {$i=1,...,M\%3$}
    \STATE generate index pool $\mathcal{I}$
    \STATE randomly choose $B_1\in\mathcal{B}$, $B_2\in\mathcal{B}$, $B_3\in\mathcal{B}$
    \STATE randomly choose $(m, n) \in \mathcal{I}$
    \STATE select $(F_1, \alpha_1) = (F_{m}, \alpha_{m}$)
    \STATE select $(F_2, \alpha_2) = (F_{n}, \alpha_{n}$)
    \STATE $S_1\leftarrow \mathcal{COMP}(F_1,\alpha_1,B_1)$ \textcolor{mygreen}{// the 1-st sample}
    \STATE $S_2\leftarrow \mathcal{COMP}(F_2,\alpha_2,B_2)$ \textcolor{mygreen}{// the 2-nd sample}
    \STATE \textcolor{mygreen}{// foreground combination}
    \STATE $(F_{\mathrm{new}},\alpha_{\mathrm{new}})\leftarrow \mathcal{RCF}(F_1,\alpha_1,F_2,\alpha_2)$
    \STATE $S_3\leftarrow \mathcal{COMP}(F_{\mathrm{new}},\alpha_{\mathrm{new}},B_3)$ \textcolor{mygreen}{// the 3-rd sample}
    \STATE $\mathcal{S}\leftarrow \mathcal{S} \cup \{S_1,S_2,S_3\}$
    \textcolor{mygreen}{// include $3$ samples into $\mathcal{S}$}
\ENDFOR
\end{algorithmic}
\end{algorithm}

\subsection{Quadruplet-style Composition}
The quadruplet-style composition is summarized in Algorithm~\ref{alg:quad-style}. Compared with triplet-style composition, the only difference is that quadruplet-style composition generates a twin foreground for the combined foreground.

\begin{algorithm}[h]\small
\caption{Quadruplet-Style Composition} 
\label{alg:quad-style} 
\begin{algorithmic}
\REQUIRE Foreground set $\mathcal{F}$, alpha set $\mathcal{A}$, and background set $\mathcal{B}$
\ENSURE Sample set $\mathcal{S}$ with $M$ samples
\STATE Initialize $\mathcal{S}\leftarrow\{\}$
\STATE \textcolor{mygreen}{// each iteration generates 4 samples}
\FOR {$i=1,...,M\%4$}
    \STATE generate index pool $\mathcal{I}$
    \STATE randomly choose $B_1\in\mathcal{B}$, $B_2\in\mathcal{B}$, $B_3\in\mathcal{B}$, $B_4\in\mathcal{B}$
    \STATE randomly choose $(m, n) \in \mathcal{I}$
    \STATE select $(F_1, \alpha_1) = (F_{m}, \alpha_{m}$)
    \STATE select $(F_2, \alpha_2) = (F_{n}, \alpha_{n}$)
    \STATE $S_1\leftarrow \mathcal{COMP}(F_1,\alpha_1,B_1)$ \textcolor{mygreen}{// the 1-st sample}
    \STATE $S_2\leftarrow \mathcal{COMP}(F_2,\alpha_2,B_2)$ \textcolor{mygreen}{// the 2-nd sample}
    \STATE \textcolor{mygreen}{// generate twins}
    \STATE $(F_{\mathrm{new1}},\alpha_{\mathrm{new1}})\leftarrow \mathcal{RCF}(F_1,\alpha_1,F_2,\alpha_2)$
    \STATE $(F_{\mathrm{new2}},\alpha_{\mathrm{new2}})\leftarrow \mathcal{RCF}(F_2,\alpha_2,F_1,\alpha_1)$
    \STATE $S_3\leftarrow \mathcal{COMP}(F_{\mathrm{new1}},\alpha_{\mathrm{new1}},B_3)$
    \textcolor{mygreen}{// the 3-rd sample}
    \STATE $S_4\leftarrow \mathcal{COMP}(F_{\mathrm{new2}},\alpha_{\mathrm{new2}},B_4)$
    \textcolor{mygreen}{// the 4-th sample}
    \STATE $\mathcal{S}\leftarrow \mathcal{S} \cup \{S_1,S_2,S_3,S_4\}$
    \textcolor{mygreen}{// include $4$ samples into $\mathcal{S}$}
\ENDFOR
\end{algorithmic}
\end{algorithm}

\begin{figure}[!t]
  \centering
  \includegraphics[width=\linewidth]{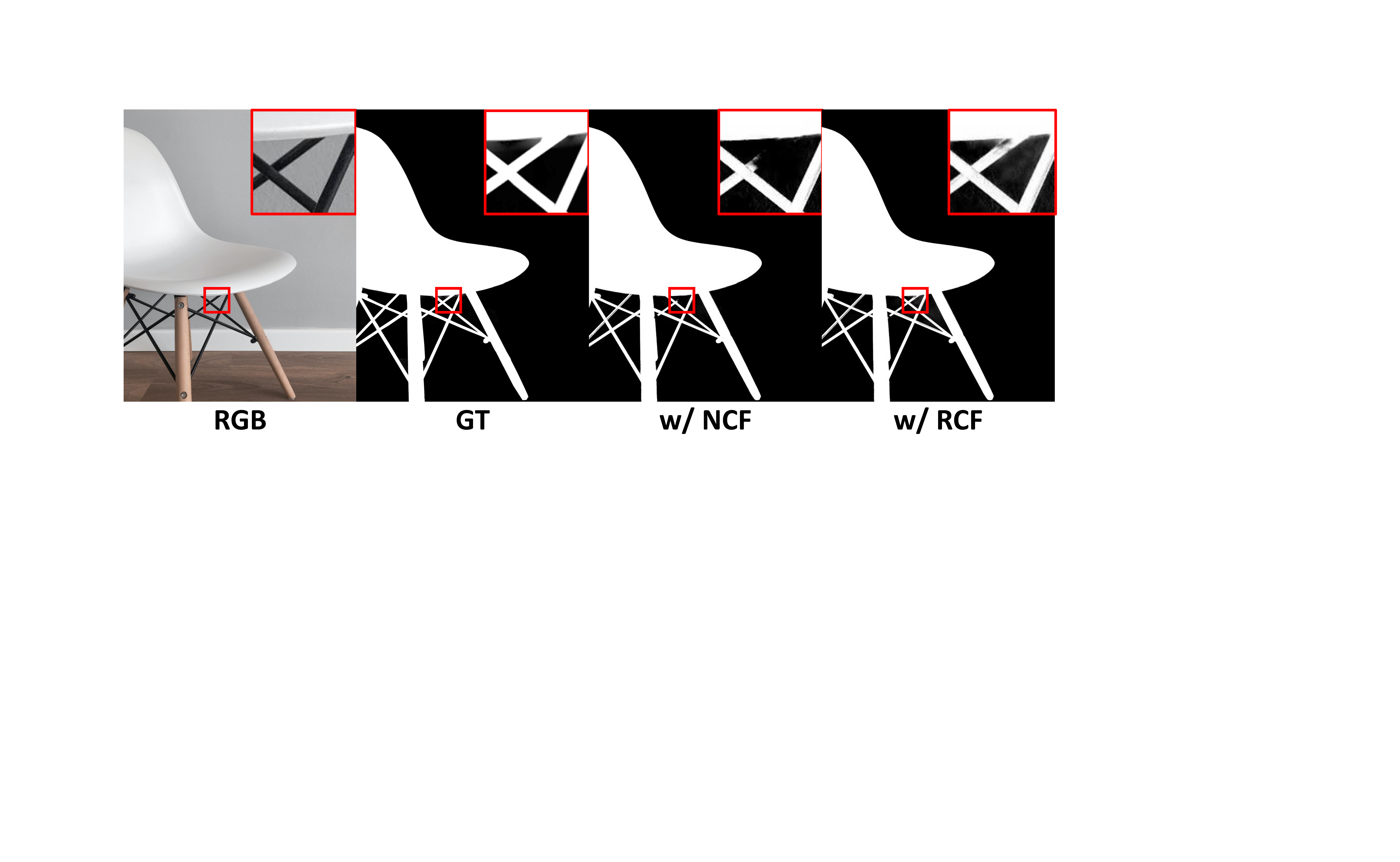}
  \caption{Visual results on real-world images. 
  From left to right: input image, GT alpha matte, predictions of
  A2U~\cite{A2U} trained with $\mathcal{NCF}$ and our $\mathcal{RCF}$, respectively.}
  \label{fig:performanceNCFRCF}
\end{figure}

\section{Visual Results of Different Foreground Combination Operators}
Here we illustrate the visual results of two foreground combination operators: $\mathcal{NCF}$ and $\mathcal{RCF}$. Visual results are depicted in Fig.~\ref{fig:performanceNCFRCF}. One can see that the chair support is missed by $\mathcal{NCF}$, whereas $\mathcal{RCF}$ can smoothly and reliably recover the details. 
We assume that the artifacts generated by $\mathcal{NCF}$ 
would affect the opacity. Due to the mismatched supervision, the network may fail to recover some simple contours.

\begin{figure*}[h]
  \centering
  \includegraphics[width=\linewidth]{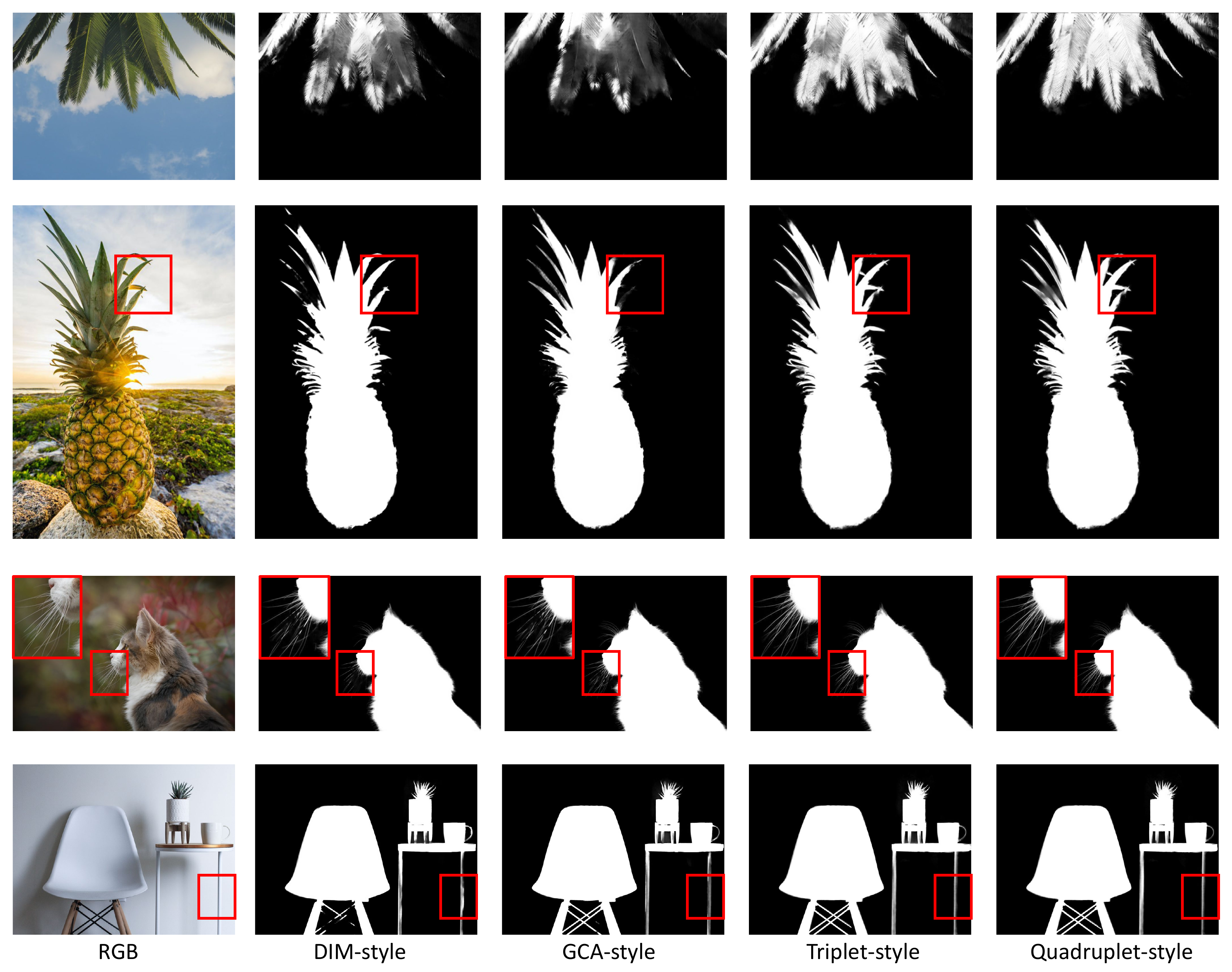}
  \caption{Visual results on real-world images with A2U as baseline.
  }
  \label{fig:VS}
  \vspace{-5pt}
\end{figure*}

\section{More Experimental Results}
Here we report more ablation studies and full experimental results.
\subsection{Comparison of Foreground Combination Operator}
In addition to the evaluation on Composition-1k~\cite{DIM}, we also conduct experiments on AIM-500~\cite{AIM} to test the robustness of $\mathcal{RCF}$ on real-world dataset.

\begin{table}[!t]\footnotesize
\centering
\addtolength{\tabcolsep}{-1
pt}
\begin{tabular}{@{}lcccccc@{}}
\toprule
 & \multirow{2}{*}{$\mathcal{NCF}$} & \multirow{2}{*}{$\mathcal{RCF}$}& \multicolumn{4}{c}{Compostion-1k} \\
 & && SAD & MSE & GRAD & CONN \\
 \midrule
A2U-quad & \Checkmark &  & 32.70 & \textbf{0.0066} & 16.24 & 30.29 \\
A2U-quad &  & \Checkmark & \textbf{30.47} & 0.0071 & \textbf{15.30} & \textbf{27.27} \\
Index-gca & \Checkmark &  &45.19  &\textbf{0.0130}  &23.62  &44.68  \\
Index-gca &  & \Checkmark & \textbf{44.03} & 0.0135 & \textbf{24.76} & \textbf{42.89}\\
\midrule
 & \multirow{2}{*}{$\mathcal{NCF}$} & \multirow{2}{*}{$\mathcal{RCF}$}&\multicolumn{4}{c}{AIM-500} \\
 &  &  & SAD & MSE & GRAD & CONN \\
\midrule
A2U-quad & \Checkmark &  & 29.21 & 0.0293 & 22.09 & 29.72 \\
A2U-quad &  & \Checkmark & \textbf{28.03} & \textbf{0.0272} & \textbf{21.21} & \textbf{28.47}\\
Index-gca & \Checkmark &  &37.99  &0.0508  &32.60  &37.82  \\
Index-gca &  & \Checkmark & \textbf{33.90} & \textbf{0.0431} & \textbf{29.96} & \textbf{33.63}\\

\bottomrule
\end{tabular}
\caption{Comparison of foreground combination operator on the Composition-1k test set and the AIM-500. A2U-quad denotes the A2U baseline with the quadruplet-style composition, and Index-gca represents the IndexNet baseline with the GCA-style composition.}
\label{tab:foreground-combination}
\end{table}

\subsection{Shuffled Samples vs. Ordered Samples}
Here we conduct an interesting investigation on the trade-off between foreground diversity and sample link. In our view, more foreground diversity represents more foreground-independent samples in a batch at training. In contrast, a deeper sample link corresponds to more foreground-related samples.
If the samples are ordered, relevant samples, \textit{i.e.}, new samples with two source foregrounds, and with their combined one(s), would be placed in the same batch. This operation introduces deeper sample link into the training process. On the contrary, shuffling enables diverse samples in a batch and establishes a soft sample link in a training epoch.
In Table~\ref{tab:distribution}, we find that ordered samples deteriorate the performance, indicating that excessively tight link among samples is counterproductive and obstructs the learning of inter-class difference.
Shuffling samples prevents the network from over-focusing on redundant features of the relevant samples. In this case, the proposed composition styles
can keep a good balance between diverse information in each batch and a global sample link
in the sample set, therefore leading to better performance.

\begin{table}[!t]\small
    \addtolength{\tabcolsep}{0pt}
    \renewcommand\arraystretch{1.1}
    \centering
    \begin{tabular}{@{}lcccccc@{}}
        \toprule
          & \multirow{2}{*}{shuffle} &  \multicolumn{4}{c}{composition-1k}               \\
          && SAD & MSE & GRAD & CONN \\
        \midrule
        triplet& \multirow{2}{*}{\Checkmark}  
        &31.52 &0.0072 &15.89 &28.47 \\
        
        quadruplet &  &30.47 &0.0071 &15.30 &27.27 \\
        \hline
        triplet& 
        &32.40 &0.0075 &16.59 &29.51 \\
        quadruplet &  &31.56 &0.0071 &16.40 &28.45\\
        \bottomrule
    \end{tabular}
    \caption{Shuffled samples vs. ordered samples on A2U.}
    \vspace{-10pt}
    \label{tab:distribution}
\end{table}

\section{Visual Results on Real-world Images}
Visual results on real-world images are present in Fig.~\ref{fig:VS}, where A2U~\cite{A2U} is employed as the baseline. Compared with DIM-style composition and GCA-style composition, our triplet-/quadruplet-style composition gives more robustness to the matting model.

\end{document}